\documentclass{bmvc2k}

\makeatletter
\renewcommand{\bmv@RenderAuthorMail}[1]{%
    \small\textcolor{bmv@sectioncolor}{#1}}
\makeatother

\usepackage{graphicx}
\usepackage{amsmath}
\usepackage{amssymb}
\usepackage{bm}
\usepackage{booktabs}
\usepackage{multirow}
\usepackage{wrapfig}
\usepackage{xcolor}
\usepackage{pifont}
\usepackage{placeins}

\newcommand{\cmark}{\ding{51}}
\newcommand{\xmark}{\ding{55}}
\newcommand{\flex}{FlexSplat}

\title{\flex: Flexible Feed-Forward 3D Gaussian Splatting without Point Cloud Correspondence}

\addauthor{Amir Sabbaghziarani}
{\shortstack[l]{%
\bmvaUrl{asabbaghziarani1@gsu.edu}\\
\bmvaUrl{asabbagh7@gatech.edu}}}
{1,2}

\addauthor{Hanting Ye}
{\bmvaUrl{hanting.ye@duke.edu}}
{3}

\addauthor{Maria Gorlatova}
{\bmvaUrl{maria.gorlatova@duke.edu}}
{3}

\addauthor{Yi Ding}
{\bmvaUrl{yding@utk.edu}}
{4}

\addinstitution{
Tri-Institutional Georgia Institute of Technology/Georgia State University/Emory University
Center for Translational Research in Data Science and Neuroimaging (TReNDS)\\
Atlanta, GA, USA
}

\addinstitution{
Georgia State University\\
Atlanta, GA, USA
}

\addinstitution{
Duke University\\
Durham, NC, USA
}

\addinstitution{
University of Tennessee, Knoxville\\
Knoxville, TN, USA
}

\runninghead{Sabbaghziarani et al.}{\flex: Correspondence-Free 3DGS}

\def\eg{\emph{e.g}\bmvaOneDot}

\begin{document}

\maketitle

\begin{abstract}
We present \flex, a feed-forward framework for novel view synthesis (NVS) from
\emph{uncalibrated}, object-centric multi-view image collections. A recent line
of query-based methods reconstructs a compact set of 3D Gaussians by treating
them as transformer queries that are refined with multi-view deformable
attention; these methods, however, assume that camera poses are given. \flex\
removes this assumption: a geometry transformer is trained \emph{jointly} with
the Gaussian decoder to predict per-image camera parameters and depth, which
in turn ground a depth-guided Gaussian parameterization and a multi-view
deformable cross-attention that aggregates evidence across all input views into
a single, view-consistent set of primitives. An uncertainty-weighted
depth-consistency objective lets the jointly trained geometry adapt to the
reconstruction task, while the cross-view consensus formed during decoding
absorbs the residual error of the estimated cameras and depth. The
representation uses a compact Gaussian budget that is decoupled from the input
resolution --- unlike pixel-aligned methods, the primitive count does not grow
with the image grid --- and is not dictated by the number of views. On ShapeNet-SRN and Google Scanned Objects (GSO), \flex\ matches or
approaches posed state-of-the-art reconstructors while requiring \emph{neither}
camera poses \emph{nor} ground-truth depth, and matches the best perceptual
(LPIPS) quality among the compared methods on GSO. Our results indicate that a
jointly trained geometry front-end is sufficient to bring calibration-free
operation to query-based Gaussian reconstruction while staying within $0.7$ dB
PSNR of posed methods and matching their perceptual quality. The code is available at \url{https://github.com/amir-sbg/FlexSplat}.
\end{abstract}

\section{Introduction}
\label{sec:intro}

Novel view synthesis (NVS) aims to render a scene from unseen viewpoints given a
set of input images. Neural Radiance Fields (NeRF)~\cite{mildenhall2020nerf} and
its variants~\cite{barron2022mipnerf360,muller2022instantngp} achieve striking
photorealism, and 3D Gaussian Splatting (3D-GS)~\cite{kerbl20233dgs} adds
real-time rendering through explicit anisotropic primitives. Both, however,
require accurate camera poses from Structure-from-Motion~\cite{schonberger2016sfm}
and per-scene optimization that takes minutes to hours. Feed-forward
reconstructors~\cite{charatan2024pixelsplat,chen2024mvsplat,SplatterImage} remove
the per-scene optimization by predicting a 3D representation directly from
images, but most of them still place Gaussians through an \emph{explicit
correspondence}. Two correspondence paradigms dominate. In the
\emph{pixel-aligned} paradigm~\cite{SplatterImage,charatan2024pixelsplat,a_pixel_more_Gau},
each pixel of each input view is mapped to one Gaussian, so the number of
primitives grows with the image resolution and the number of views. In the
\emph{point-based} paradigm~\cite{zou2023triplane,xu2024grm}, Gaussians are
anchored to dense 3D points or triplane features predicted by a multi-stage
geometry pipeline. We use these two terms throughout in this precise sense: a
pixel-aligned primitive is tied to an image location, a point-based primitive is
tied to a predicted 3D location.

Both paradigms share a structural cost. Tying primitives to pixels or points
forces the model to spend Gaussians on the input parameterization rather than on
the object, producing redundant, overlapping primitives that overfit appearance,
inflate memory, and blur regions where views disagree~\cite{wu2024leangaussian,unigs}.
A recent and effective alternative breaks the correspondence entirely: Gaussians
are modelled as \emph{learnable transformer queries}, each query is a 3D
ellipsoid whose centre is a reference point projected onto image features, and a
deformable decoder refines the queries layer by layer.
LeanGaussian~\cite{wu2024leangaussian} introduced this idea for the
\emph{single-view} case, and UniGS~\cite{unigs} extended it to multiple
\emph{posed} views through a multi-view deformable cross-attention that
aggregates a single, ``unitary'' set of Gaussians across views. These methods
produce compact, high-quality reconstructions, but they all assume that camera
poses (and often depth) are available at input time --- precisely the
preprocessing that feed-forward NVS set out to avoid.

\flex\ closes this gap for the object-centric setting. Our scope is deliberate:
we target uncalibrated reconstruction of \emph{objects} --- the regime of asset
creation, image-to-3D generation, and robotic object manipulation --- rather than
large, background-heavy scenes. Within this scope we keep the query-based
unitary-Gaussian decoder of LeanGaussian and UniGS, but we replace the
assumption of known geometry with a geometry transformer
(VGGT~\cite{wang2025vggt}) that is trained \emph{jointly} with the decoder and
predicts per-image cameras and depth. The decoder is then grounded in this
learned geometry in three ways: the predicted depth seeds a depth-guided
parameterization of each Gaussian centre; fused depth-and-appearance features
serve as keys and values for the multi-view deformable attention; and an
uncertainty-weighted depth-consistency loss aligns the rendered geometry with
the transformer's depth. Because every Gaussian aggregates evidence from all
views and the per-view contributions are reconciled in world space, the decoder
forms a cross-view consensus that is tolerant to the imperfect poses and depth it
is given --- a property we exploit rather than assume away. The Gaussian budget is
decoupled from resolution and not dictated by the number of views.

We make the following contributions:
\begin{itemize}
\item We present \flex, a feed-forward framework that brings query-based,
correspondence-free Gaussian reconstruction to the \emph{uncalibrated}
object-level setting by training a geometry transformer jointly with the
Gaussian decoder, removing the posed-input requirement of prior query-based
methods~\cite{wu2024leangaussian,unigs}.
\item We introduce a geometry-grounded decoder: a depth-guided Gaussian
parameterization and a multi-view deformable cross-attention over fused
depth/appearance features, supervised by an uncertainty-weighted
depth-consistency objective. The resulting cross-view consensus is empirically
robust to the noise of jointly estimated cameras and depth.
\item We show that \flex\ attains a compact Gaussian representation whose budget
is decoupled from input resolution and not dictated by the number of
views, that matches posed state-of-the-art reconstructors (\eg UniGS) and
matches the best perceptual quality on GSO \emph{without} any camera or depth
ground truth.
\end{itemize}

\section{Related Work}
\label{sec:related}

\paragraph{Query-based, correspondence-free Gaussian reconstruction.}
The closest line of work to ours abandons pixel/point correspondence and treats
3D Gaussians as transformer queries. LeanGaussian~\cite{wu2024leangaussian}
defines each query as a Gaussian ellipsoid whose centre is a 3D reference point
projected onto image features for deformable attention, and refines the queries
in a single-view decoder. UniGS~\cite{unigs} generalises this to multiple
\emph{posed} views with a multi-view deformable cross-attention (MVDFA) that
updates one unitary set of Gaussians shared across views, reducing the
``ghosting'' and redundancy of per-pixel methods. C3G~\cite{c3g} pushes the same
query-based idea towards extreme compactness, reconstructing scenes from unposed
images with only $\sim$2K Gaussians. GeoLRM~\cite{geolrm} likewise uses
deformable cross-attention from 3D anchor points to image features. \flex\
inherits the decoder design of this family but differs in what it assumes about
geometry: rather than consuming given poses (UniGS, GeoLRM) or operating purely
in image space, it integrates a \emph{jointly trained} geometry transformer that
supplies cameras, depth, and depth-aware features, enabling calibration-free
object reconstruction.

\paragraph{Feed-forward object reconstruction.}
Large reconstruction models predict 3D assets from one or a few images.
Splatter Image~\cite{SplatterImage} and its hierarchical
extension~\cite{a_pixel_more_Gau} regress pixel-aligned Gaussians;
LGM~\cite{LGM}, GRM~\cite{xu2024grm}, GS-LRM~\cite{gslrm}, and
MVGamba~\cite{mvgamba} reconstruct objects from posed multi-view images;
InstantMesh~\cite{instantmesh} and DreamGaussian~\cite{tang2023dreamgaussian}
target mesh/3D content generation. All of these require posed inputs and/or
multi-view diffusion to synthesise the views, and most scale their primitive
count with the input. \flex\ instead operates on \emph{uncalibrated} captures and
keeps a compact Gaussian budget.

\paragraph{Pose-free and scene-level synthesis.}
A parallel and very active line targets large, background-heavy \emph{scenes}
from unposed images. AnySplat~\cite{anysplat} distils geometry priors from
VGGT~\cite{wang2025vggt} and prunes Gaussians by voxelization;
NoPoSplat~\cite{noposplat} predicts Gaussians in a canonical frame from unposed
pairs; DepthSplat~\cite{depthsplat} couples Gaussian prediction with monocular
depth; MVSplat~\cite{chen2024mvsplat} and PF3plat~\cite{hong2024pf3plat} build
cost volumes or use depth/correspondence priors; and WorldMirror~\cite{worldmirror}
unifies many geometric outputs under flexible prior prompting. These methods are
trained and evaluated on scene benchmarks (\eg RealEstate10K, DL3DV, ACID) whose
imagery, scale, and pixel-aligned representations differ fundamentally from the
object-centric, masked-object setting we study; their public models target
backgrounds and large baselines rather than single objects. We therefore discuss
them as context but do not claim head-to-head comparison on object benchmarks,
which would not be a like-for-like evaluation. The same observation has been
raised for object datasets such as ShapeNet-SRN and GSO. \flex\ occupies the
object-level, uncalibrated niche between posed object reconstructors (UniGS, LGM,
GRM) and scene-level pose-free models (AnySplat, NoPoSplat).

\paragraph{Geometry foundation models and deformable attention.}
DUSt3R~\cite{wang2024dust3r}, Fast3R~\cite{yang2025fast3r}, and
VGGT~\cite{wang2025vggt} predict dense geometry and cameras directly from images
and have become strong priors for downstream 3D tasks. We adopt VGGT as the
geometry front-end but, unlike methods that consume it frozen as a fixed prior,
we train it jointly with the decoder under the rendering and depth-consistency
objectives. Deformable attention~\cite{zhu2021deformabledetr} originates in 2D
detection, where features are sampled around reference points; we use its
multi-view 3D extension as in~\cite{unigs,wu2024leangaussian}.

\section{Method}
\label{sec:method}

\begin{figure*}
\centering
\includegraphics[width=\textwidth]{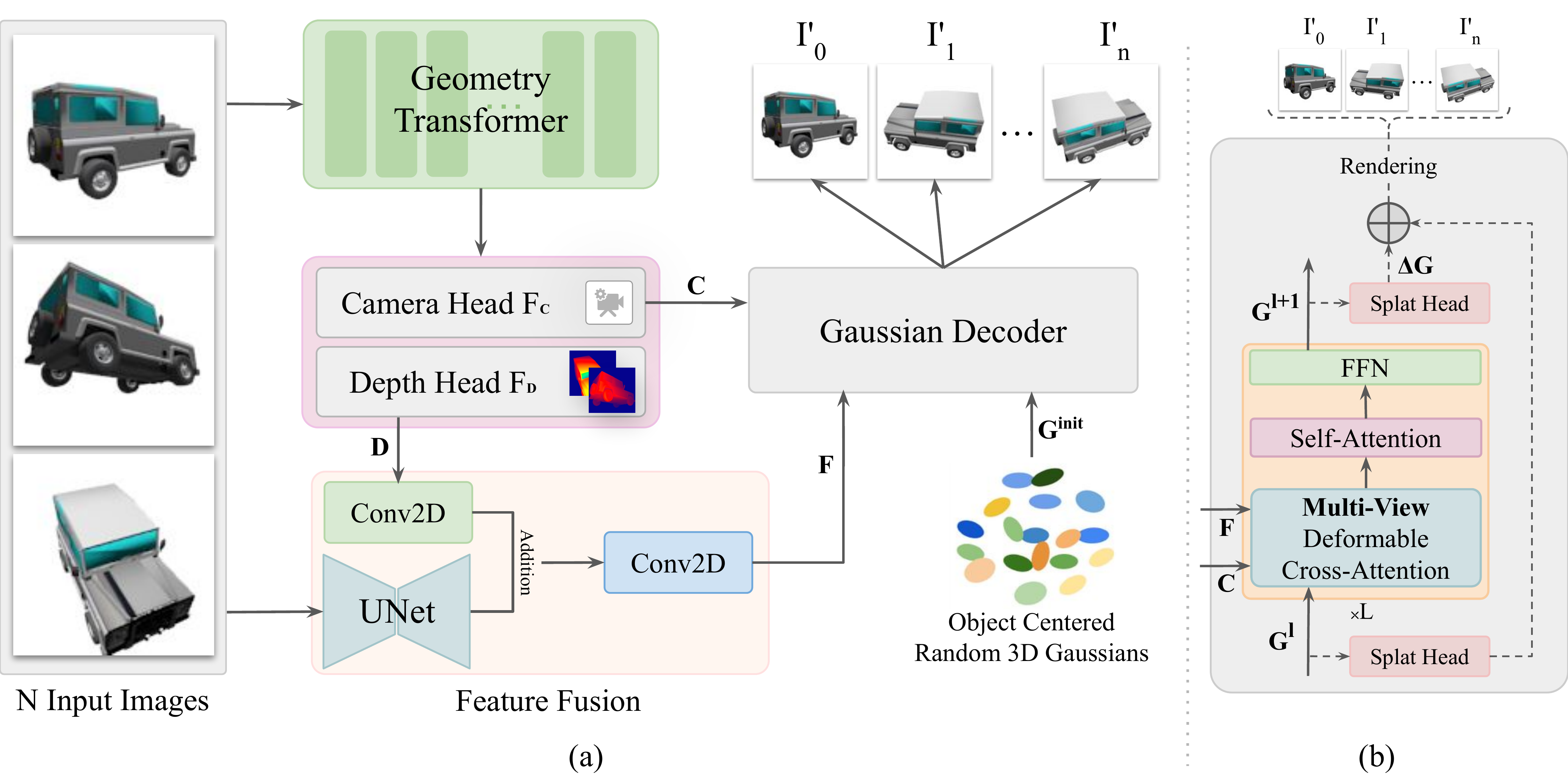}
\caption{\textbf{Overview of \flex.}
(a) Given $N$ ($N \geq 1$) uncalibrated input images $\{I_i\}_{i=1}^{N}$, a
pretrained VGGT~\cite{wang2025vggt}, trained jointly with the decoder, predicts
camera parameters and depth maps via the camera head $F_C$ and depth head $F_D$.
The predicted depth maps are fused with UNet-extracted pixel-aligned features to
obtain depth-aware feature maps $F$.
A set of random 3D Gaussians $G^{init}$, initialized around the scene center,
serves as queries for the Gaussian decoder.
(b) The decoder, with $L$ iterative layers, refines Gaussians through multi-view
deformable cross-attention, self-attention, and a feed-forward network (FFN).
Reference points are projected onto multi-view feature planes at each layer,
while the Splat Head renders outputs for supervision.
$G^l$ denotes the Gaussian queries at layer $l$. $\dashrightarrow$ indicates
steps used during training only.}
\label{fig:flexsplat_architecture}
\end{figure*}

\flex\ maps a set of uncalibrated object images to a single, compact set of 3D
Gaussians and renders them to novel views. We review 3D-GS
(Sec.~\ref{sec:bg}), state the problem (Sec.~\ref{sec:problem}), describe the
architecture (Sec.~\ref{sec:arch}), analyse the compact representation
(Sec.~\ref{sec:budget}), and give the training objective (Sec.~\ref{sec:obj}).
The overall pipeline is shown in Figure~\ref{fig:flexsplat_architecture}.

\subsection{Background: 3D Gaussian Splatting}
\label{sec:bg}
A scene is represented by anisotropic Gaussian ellipsoids, each with a mean
$\bm{\mu}$ and covariance $\bm{\Sigma}=\mathbf{R}\mathbf{S}\mathbf{S}^{\top}\mathbf{R}^{\top}$,
decomposed into a quaternion rotation $\mathbf{R}$ and a diagonal scale
$\mathbf{S}$ for numerical stability~\cite{kerbl20233dgs}. Each Gaussian carries
an opacity $\sigma\in[0,1]$ and spherical-harmonic colour. As an extension of
point-based splatting~\cite{zwicker2001ewa}, the pixel colour is obtained by
alpha-blending the projected Gaussians along the ray, which is differentiable
and renders in real time.

\subsection{Problem Formulation}
\label{sec:problem}
Given $N$ RGB images $\{I_i\in\mathbb{R}^{H\times W\times3}\}_{i=1}^{N}$ of an
object \emph{without} known calibration, we learn a function that predicts (i)
per-image geometry --- camera parameters $\mathbf{c}_i=\{\mathbf{R}_i,\mathbf{t}_i\}$
and a dense depth map $d_i\in\mathbb{R}^{H\times W}$ --- and (ii) a scene
representation $\mathbf{G}=\mathcal{G}_{\Phi}(\{I_i\}_{i=1}^{N})=\{g_n\}_{n=1}^{N_G}$
of $N_G$ Gaussians, each
$g_n=\{\bm{\mu}_n,\mathbf{R}_n,\mathbf{S}_n,\sigma_n,\mathbf{SH}_n\}$. A novel
view at camera $\pi_{j'}$ is rendered by the differentiable splatter
$\mathcal{R}$:
\begin{equation}
I_{j'}^{\text{pred}}=\mathcal{R}\big(\mathcal{G}_{\Phi}(\{I_i\}_{i=1}^{N}),\,\pi_{j'}\big),
\end{equation}
and supervised against the ground-truth image $I_{j'}$. Crucially, all of
$\mathbf{c}_i$, $d_i$, and $\pi_{j'}$ are \emph{predicted}, not given.

\paragraph{Depth-guided mean parameterization.}
Regressing absolute centres $\bm{\mu}_n$ directly is unstable. Following the
depth grounding of~\cite{SplatterImage}, we tie each centre to the predicted
depth and a learnable offset. Let $(u_{1i},u_{2i})$ denote the calibrated
(normalised) image coordinates of the corresponding location in view $i$, and
$(\Delta_{x_i},\Delta_{y_i},\Delta_{z_i})$ a learnable offset; the centre
expressed in the camera frame of view $i$ is the back-projection
\begin{equation}\label{eq:center}
\bm{\mu}_{n,i}=
\big[\,u_{1i}d_i+\Delta_{x_i},\;\; u_{2i}d_i+\Delta_{y_i},\;\; d_i+\Delta_{z_i}\,\big]^{\top}.
\end{equation}
Grounding the centre in the predicted depth stabilises convergence while leaving
room for cross-view refinement; because the depth itself is trained
(Sec.~\ref{sec:obj}), this prior adapts to the reconstruction task rather than
fixing errors in place.

\subsection{Architecture}
\label{sec:arch}

\paragraph{Jointly trained geometry transformer.}
We use VGGT~\cite{wang2025vggt} with two heads: a dense-prediction
head~\cite{ranftl2021dense} producing per-pixel depth and geometry-aware
features, and a camera head regressing intrinsics and extrinsics for each image.
In contrast to prior pipelines that treat such a model as a fixed front-end, we
fine-tune VGGT jointly with the decoder, so its depth and camera predictions are
shaped by the rendering and depth-consistency objectives. This is the property
that lets the system partially self-correct: the geometry is not asked to be
right in isolation, only to be useful for the reconstruction it is supervised to
produce.

\paragraph{Dual-branch features and fusion.}
A UNet encoder--decoder~\cite{ronneberger2015unet} extracts pixel-aligned
appearance features in parallel with the geometry branch. Following the
FPN-style fusion of~\cite{wu2024leangaussian,lin2017fpn}, depth features are
added to the UNet features and refined by a convolution, then reduced by a
lightweight convolution for efficiency. The fused multi-view features $F$ serve
as the keys and values of the deformable cross-attention. A reconstruction loss
on the UNet features enforces pixel-wise consistency between input and
reconstructed views.

\paragraph{Geometric initialization.}
Random Gaussian placement is unstable in the multi-view setting, where sampling a
projected centre that lands off-image yields zero features and vanishing
gradients. We therefore place all queries near the estimated scene centre,
computed as the least-squares intersection of the camera optical axes from the
camera head, and regress the initial parameters
$\mathbf{G}_{\mathrm{init}}=\mathcal{S}(\mathbf{q}_{\mathrm{init}})$ with a
splatting head using Xavier-uniform weights and fixed attribute biases.


\paragraph{Multi-view deformable cross-attention.}
\label{sec:mvattn}
\begin{wrapfigure}{l}{0.4\textwidth}
    \centering
    \includegraphics[width=0.4\textwidth]{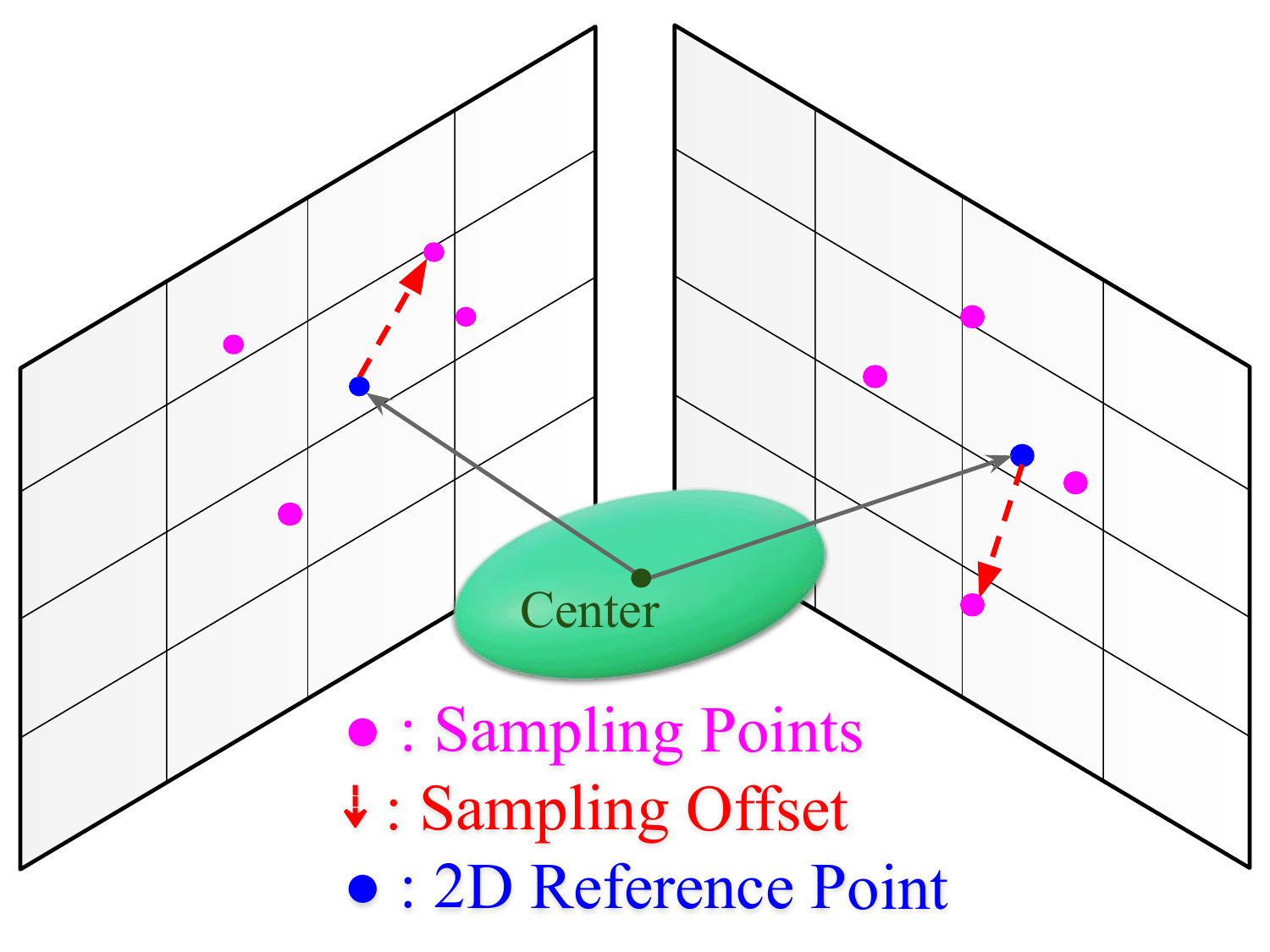}
    \caption{\textbf{Multi-view projection of Gaussian centers.}
    Each 3D Gaussian is projected onto multiple image planes, where its 2D
    reference points and learnable offsets define sampling locations for
    multi-view deformable attention.}
    \label{fig:multiview_gaussian}
\end{wrapfigure}
We adopt the multi-view deformable formulation of~\cite{unigs,wu2024leangaussian}.
Each query is a Gaussian ellipsoid with no native 2D location; we obtain one by
projecting its 3D centre into every view (Figure~\ref{fig:multiview_gaussian}). Let $\pi_i(\cdot)$ denote the
perspective projection into view $i$ (the rigid transform by
$(\mathbf{R}_i,\mathbf{t}_i)$, the intrinsics $\mathbf{K}_i$, and the division by
depth that maps to pixel coordinates). At decoder layer $l$, the centre
$\bm{\mu}_n^{l}$ yields $N$ reference points
$\{\pi_1(\bm{\mu}_n^{l}),\dots,\pi_N(\bm{\mu}_n^{l})\}$ describing how the same
primitive is seen across views. Around each reference point we sample $N_P$
locations with learned offsets and aggregate them with attention weights.
Figure~\ref{fig:multiview_gaussian} illustrates the 2D reference point,
the sampled locations, and their offsets in each view. The aggregation is:
\begin{equation}\label{eq:attn}
\mathbf{q}_{n}^{l'}=\sum_{i=1}^{N}\sum_{p=1}^{N_P}
A_{n,i,p}^{l}\;\mathrm{Bilinear}\!\big(F_i,\;\pi_i(\bm{\mu}_n^{l})+\Delta\mathbf{s}_{n,i,p}^{l}\big),
\end{equation}
where $\Delta\mathbf{s}_{n,i,p}^{l}=\mathrm{MLP}_{\text{off}}(\mathbf{q}_n^{l})$
are the learned sampling offsets and the weights
$A_{n,i,p}^{l}=\mathrm{softmax}(\mathrm{MLP}^{A}(\mathbf{q}_n^{l}))$ are
normalised jointly over all views and sampling points
($\sum_{i,p}A_{n,i,p}^{l}=1$). Because a single query attends jointly across all
$N$ views and the resulting update is reconciled in world space, each Gaussian
forms a cross-view consensus rather than committing to any one (possibly noisy)
projection --- the mechanism that absorbs moderate camera and depth error.

\paragraph{Gaussian decoder and refinement.}
\label{sec:decoder}
We learn $N$ latent queries $\mathbf{q}\in\mathbb{R}^{N\times C}$ refined over $L$
layers. Each layer applies multi-view deformable cross-attention
(Eq.~\ref{eq:attn}), self-attention across queries, and an FFN:
\begin{equation}
\mathbf{q}^{l+1}=\mathrm{FFN}\big(\mathrm{SelfAttn}(\mathrm{MV\text{-}DeformAttn}(\mathbf{q}^{l},\mathbf{F},\bm{\mu}^{l}))\big),
\end{equation}
and a splatting head predicts incremental updates
$\bm{\Delta G}^l=\mathcal{S}^l(\mathbf{q}^l)$ that are composed onto the running
estimate ($\mathbf{G}^{l+1}=\mathbf{G}^{l}\oplus\bm{\Delta G}^l$: addition for
translation/appearance, multiplication for rotation). After each layer the
refined centres are re-projected to update the reference points, and each
layer's Gaussians are rendered and supervised, giving progressive,
geometry-consistent refinement.

\subsection{A Compact, Resolution-Independent Representation}
\label{sec:budget}
A direct consequence of the query-based design is that the number of primitives
$N_G$ is a free hyper-parameter, set independently of the input rather than
dictated by it. This contrasts sharply with pixel-aligned reconstructors, whose
primitive count is $\Theta(H\cdot W\cdot N)$ and therefore grows with both
resolution and the number of views. For four input views at $256{\times}256$, a
pixel-aligned method instantiates on the order of $256^2\times4\approx2.6\times10^{5}$
Gaussians, whereas \flex\ uses $2{\times}10^{4}$ --- about an order of magnitude
fewer --- and, crucially, this budget is independent of image resolution. The
budget is also not tied to the number of views: we set it to 10K/15K/20K for
1/2/4 views (Sec.~\ref{sec:viewcount}), a modest scaling chosen to give more
views more capacity, though it could equally be held fixed. Even at the largest
setting the primitive count stays roughly an order of magnitude below a
pixel-aligned four-view budget. The same motivation drives recent compact
query-based work such as C3G~\cite{c3g}.

\subsection{Training Objective}
\label{sec:obj}
With predicted depth and per-pixel confidence $(D_i^{\mathrm{VGGT}},\mathcal{U}_i)$
from the geometry head, the loss combines an RGB reconstruction term on the
feature-extractor output $\hat I_i^{E}$ and every decoder layer
$\hat I_{i,l}^{D}$, a confidence-weighted depth-consistency term aligning the
rendered depth $\hat D_i$ with $D_i^{\mathrm{VGGT}}$ (the weighting down-scales
pixels the geometry head deems unreliable), and an LPIPS~\cite{zhang2018lpips}
perceptual term:
\begin{align}
\mathcal{L}
&=\frac{1}{N}\sum_{i=1}^{N}\Big(\lambda_E\,\mathcal{L}_{\mathrm{MSE}}(\hat I_i^{E},I_i)
+\sum_{l=1}^{L}\lambda_D\,\mathcal{L}_{\mathrm{MSE}}(\hat I_{i,l}^{D},I_i)\Big) \nonumber\\
&\quad+\frac{\lambda_{\mathrm{Depth}}}{N}\sum_{i=1}^{N}\big\|(\hat D_i-D_i^{\mathrm{VGGT}})\odot\mathcal{U}_i\big\|_{1}
+\frac{1}{N}\sum_{i=1}^{N}\sum_{l=1}^{L}\lambda_{\mathrm{LPIPS}}\,\mathcal{L}_{\mathrm{LPIPS}}(\hat I_{i,l}^{D},I_i).
\end{align}
Because the geometry transformer is in the gradient path, the depth-consistency
term both regularises the Gaussians and adapts the predicted geometry to the
rendering task.

\section{Experiments}
\label{sec:exp}

\subsection{Implementation details}
\label{sec:impl}
Following~\cite{SplatterImage}, each Gaussian is parameterised by 24 values: 4
for the mean (1 depth, 3 offsets), 7 for the covariance (3 scale, 4 rotation), 1
opacity, and 12 spherical-harmonic colour coefficients. The decoder uses $L{=}3$
layers, trading off memory, accuracy, and speed. We use 10K Gaussians for
single-view and 20K for four-view testing for budget parity with baselines.
Training uses Adam with learning rate $1\times10^{-4}$ on $4\times$ RTX~4090
(24\,GB) GPUs, and begins with RGB MSE and depth losses, adding LPIPS later for
perceptual quality.

\subsection{Datasets and metrics}
\label{sec:data}
For single-view reconstruction we use ShapeNet-SRN~\cite{sitzmann2019srn} (Cars,
Chairs). For multi-view training we use a subset of
Objaverse-LVIS~\cite{deitke2023objaverse} ($>$1,000 categories), sampling 1--8
views per instance, and evaluate on 250 objects from Google Scanned Objects
(GSO)~\cite{downs2022gso} with four input views. We report PSNR, SSIM, and LPIPS
between rendered and held-out target views. \flex\ uses no camera poses
\emph{at input}; for quantitative evaluation, the cameras predicted by the
geometry head are aligned to the evaluation coordinate frame by a global
similarity transform, so that each held-out target view can be rendered and
compared against its ground truth, as is standard for pose-free
NVS~\cite{noposplat}.

\subsection{Single-view reconstruction on ShapeNet-SRN}
\label{sec:single}
\begin{figure}[t]
\centering
\includegraphics[width=0.8\linewidth]{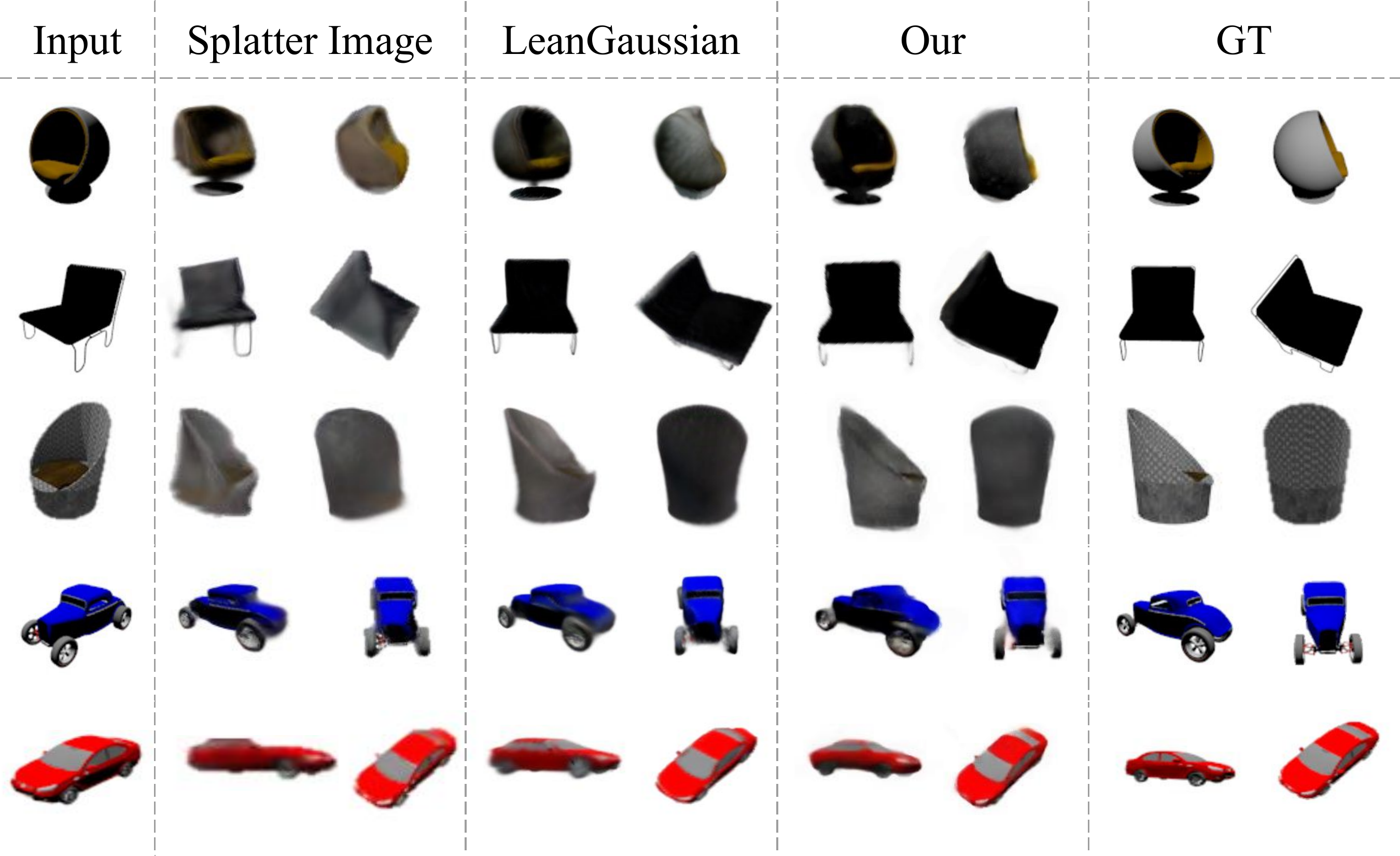}
\caption{Single-view NVS on ShapeNet-SRN. \flex\ improves geometric consistency
and structural detail over Splatter Image~\cite{SplatterImage} and
LeanGaussian~\cite{wu2024leangaussian}, with comparable colour fidelity.}
\label{fig:srn}
\end{figure}

Table~\ref{tab:srn} compares \flex\ with single-view reconstructors on
ShapeNet-SRN. \flex\ is on par with the strongest query-based baseline,
LeanGaussian~\cite{wu2024leangaussian} --- slightly ahead on Cars PSNR and within
$0.4$ dB on Chairs --- while, unlike LeanGaussian, it does not assume the input
view is the reference frame and instead infers the camera. Qualitatively
(Figure~\ref{fig:srn}), \flex\ recovers sharper boundaries and thin structures,
indicating that the multi-view aggregation benefits even the single-view regime.

\begin{table}[t]
\centering\footnotesize
\setlength{\tabcolsep}{4pt}
\caption{Single-view NVS on ShapeNet-SRN, averaged per category. Best per column
in \textbf{bold}.}
\label{tab:srn}
\begin{tabular}{lcccccc}
\toprule
& \multicolumn{3}{c}{Chairs} & \multicolumn{3}{c}{Cars}\\
\cmidrule(lr){2-4}\cmidrule(lr){5-7}
Method & PSNR$\uparrow$ & SSIM$\uparrow$ & LPIPS$\downarrow$ & PSNR$\uparrow$ & SSIM$\uparrow$ & LPIPS$\downarrow$\\
\midrule
SRN~\cite{sitzmann2019srn} & 22.89 & 0.890 & 0.104 & 22.25 & 0.880 & 0.129\\
CodeNeRF~\cite{codenerf} & 23.66 & 0.900 & 0.166 & 23.80 & 0.910 & 0.128\\
ViewsetDiff (w/o depth)~\cite{viewset_diff} & 24.16 & 0.910 & 0.088 & 23.21 & 0.900 & 0.116\\
PixelNeRF~\cite{pixelnerf} & 23.72 & 0.900 & 0.128 & 23.17 & 0.890 & 0.146\\
NeRFDiff (w/o NGD)~\cite{nerfdiff} & 24.80 & 0.930 & 0.070 & 23.95 & 0.920 & 0.092\\
Splatter Image~\cite{SplatterImage} & 24.43 & 0.930 & 0.067 & 24.00 & 0.920 & 0.078\\
Hierarchical SI~\cite{a_pixel_more_Gau} & 25.43 & 0.940 & 0.066 & 24.18 & 0.920 & 0.087\\
LeanGaussian~\cite{wu2024leangaussian} & \textbf{25.87} & \textbf{0.950} & \textbf{0.065} & 25.00 & \textbf{0.930} & \textbf{0.075}\\
\midrule
\flex\ (Ours) & 25.48 & \textbf{0.950} & 0.066 & \textbf{25.09} & 0.920 & 0.079\\
\bottomrule
\end{tabular}
\end{table}

\subsection{Multi-view reconstruction on GSO}
\label{sec:multi}
\begin{figure}[t]
\centering
\includegraphics[width=0.95\linewidth]{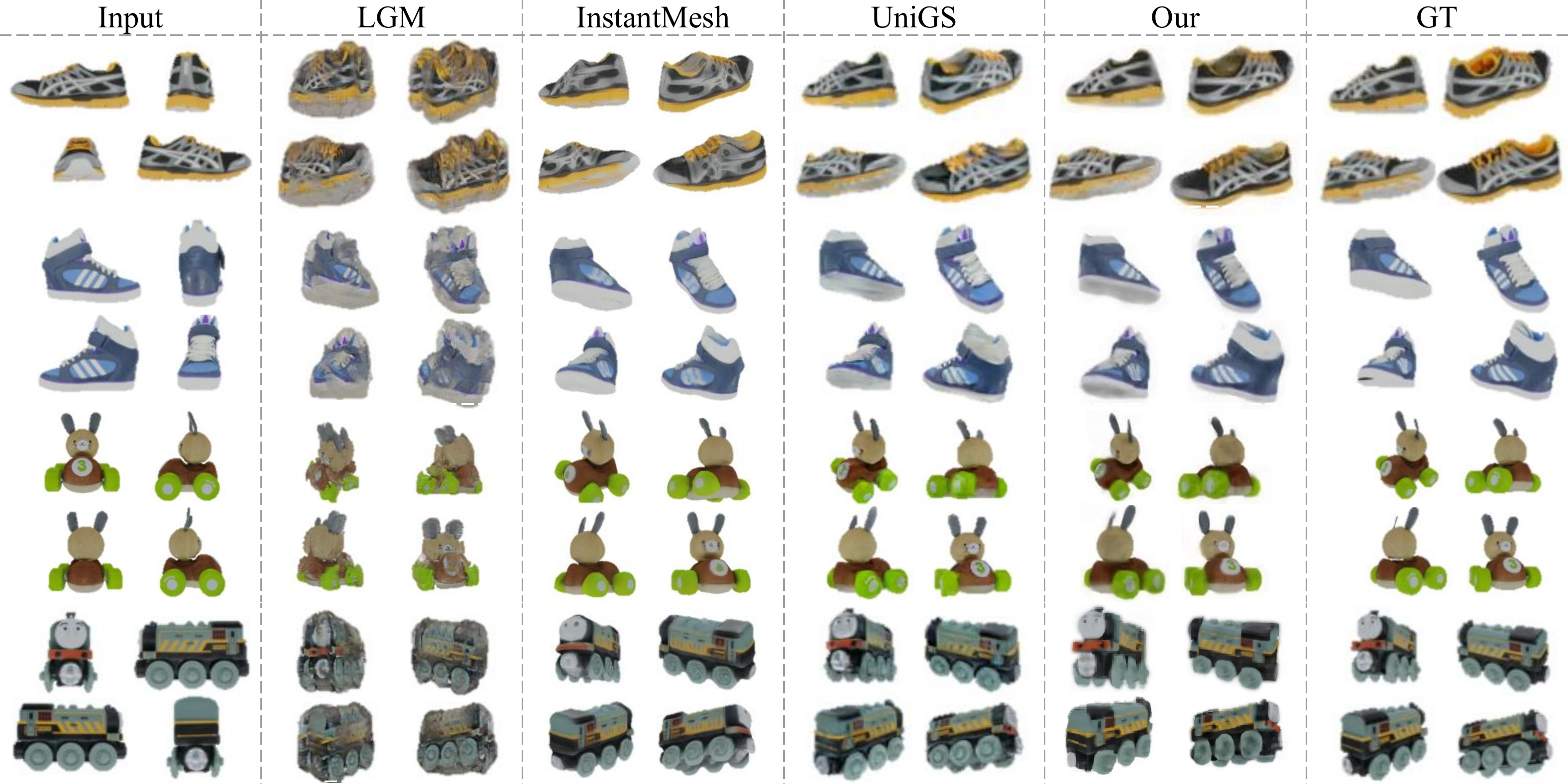}
\caption{Four-view NVS on GSO. \flex\ produces consistent geometry and texture
without any known camera parameters.}
\label{fig:gso}
\end{figure}

Table~\ref{tab:gso} reports four-view GSO results and marks which methods require
posed inputs. Every baseline is posed; \flex\ is the only pose-free method, and
like the others it uses no ground-truth depth (its depth is predicted by the
geometry head). Under this handicap \flex\ rivals the posed state of the art: it
is within $0.7$ dB PSNR and $0.012$ SSIM of UniGS~\cite{unigs} --- the method
whose query-based decoder is closest to ours but which is given the cameras ---
and it matches the best LPIPS among the compared methods ($0.041$ vs.\ $0.042$ for UniGS, within noise). We read this as evidence
that calibration-free operation incurs only a small fidelity cost here ($0.69$ dB
PSNR and $0.012$ SSIM below UniGS, with comparable LPIPS) while removing UniGS's
posed-input assumption. Qualitatively (Figure~\ref{fig:gso}), \flex\ avoids the
duplicate ``ghost'' Gaussians of LGM~\cite{LGM} and the missing fine structures
of InstantMesh~\cite{instantmesh}, and is visually comparable to UniGS despite
using no calibration.

\begin{table}[t]
\centering\footnotesize
\setlength{\tabcolsep}{4pt}
\caption{Four-view NVS on GSO. ``Posed'' marks methods that require ground-truth
or input camera poses; \flex\ is the only pose-free method. No listed method
requires ground-truth depth. $^{\dagger}$Splatter Image uses cameras only inside
the renderer, not as network input. ``NA'': not reported by the source. Best per
metric in \textbf{bold}.}
\label{tab:gso}
\begin{tabular}{lcccc}
\toprule
\multirow{2}{*}{Method} & \multicolumn{3}{c}{GSO (4 views)} & \multirow{2}{*}{Posed}\\
\cmidrule(lr){2-4}
& PSNR$\uparrow$ & SSIM$\uparrow$ & LPIPS$\downarrow$ & \\
\midrule
Splatter Image~\cite{SplatterImage} & 25.62 & 0.915 & 0.152 & \cmark$^{\dagger}$\\
LGM (Small)~\cite{LGM} & 17.48 & 0.783 & 0.218 & \cmark\\
LGM (Large)~\cite{LGM} & 26.25 & 0.925 & 0.054 & \cmark\\
InstantMesh~\cite{instantmesh} & 23.02 & 0.889 & 0.089 & \cmark\\
GeoLRM~\cite{geolrm} & 22.84 & 0.851 & NA & \cmark\\
MVGamba~\cite{mvgamba} & 26.25 & 0.881 & 0.069 & \cmark\\
GRM (Res-512)~\cite{xu2024grm} & 30.05 & 0.906 & 0.052 & \cmark\\
GS-LRM (Res-256)~\cite{gslrm} & 29.59 & 0.944 & 0.051 & \cmark\\
UniGS~\cite{unigs} & \textbf{30.42} & \textbf{0.961} & 0.042 & \cmark\\
\midrule
\flex\ (Ours) & 29.73 & 0.949 & \textbf{0.041} & \textbf{\xmark}\\
\bottomrule
\end{tabular}
\end{table}

\subsection{Single-view reconstruction on GSO}
\label{sec:gso_single}
Table~\ref{tab:gso_single} reports single-view GSO results following the UniGS
protocol (views for non-native-single-view baselines generated with
ImageDream~\cite{imagedream2025}). \flex\ exceeds LGM and InstantMesh in PSNR and
matches the best LPIPS, remaining competitive with UniGS in this open-domain
single-view setting as well.

\begin{table}[t]
\centering\footnotesize
\caption{Single-view reconstruction on GSO. Best per metric in \textbf{bold}.}
\label{tab:gso_single}
\begin{tabular}{lccc}
\toprule
Method & PSNR$\uparrow$ & SSIM$\uparrow$ & LPIPS$\downarrow$\\
\midrule
LGM~\cite{LGM} & 20.81 & \textbf{0.858} & 0.151\\
InstantMesh~\cite{instantmesh} & 19.47 & 0.838 & 0.184\\
UniGS~\cite{unigs} & \textbf{22.35} & 0.857 & \textbf{0.149}\\
\midrule
\flex\ (Ours) & 21.93 & 0.850 & \textbf{0.149}\\
\bottomrule
\end{tabular}
\end{table}

\subsection{Effect of the number of input views}
\label{sec:viewcount}
\begin{figure}[t]
\centering
\includegraphics[width=0.65\linewidth]{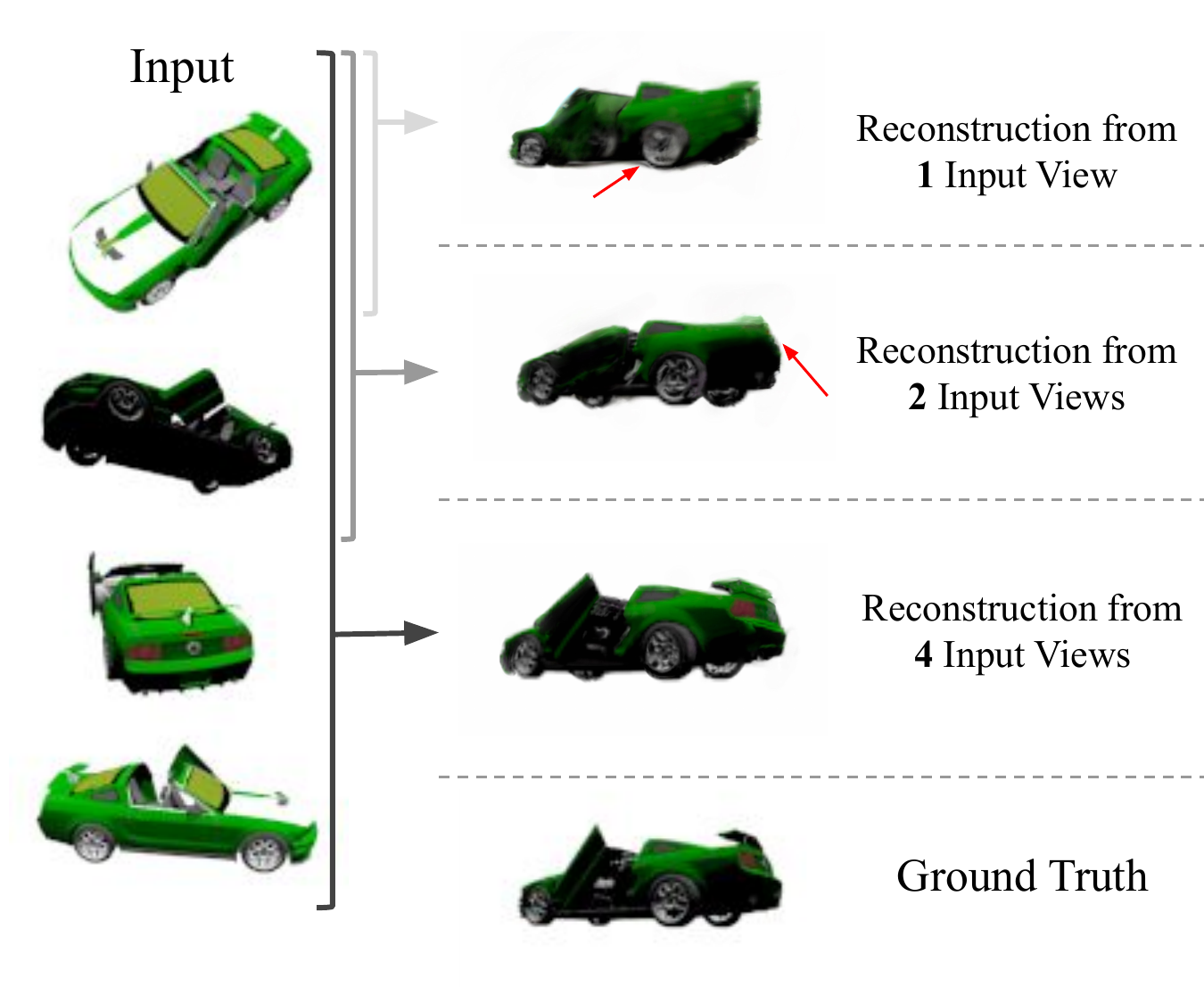}
\caption{Effect of the number of input views (1, 2, 4) for the same object.
More views give the unitary representation more cross-view evidence, improving
reconstruction. Viewpoints are randomly sampled.}
\label{fig:viewcount}
\end{figure}

Table~\ref{tab:viewcount} and Figure~\ref{fig:viewcount} quantify the effect of
the number of input views. Going from one to two views yields the largest gain
($+5.17$ dB PSNR), as the second view resolves most of the single-view
ambiguity; a fourth view adds a further $+2.63$ dB, with diminishing returns.
We scale the budget modestly with views (10K/15K/20K for 1/2/4 views) to match
the additional surface coverage; because the number of Gaussian queries is a
free hyper-parameter independent of the view count, no retraining is needed to
change the input count.

\begin{table}[t]
\centering\footnotesize
\caption{Effect of the number of input views. The same 250 GSO objects and
identical held-out target views are used across all view counts; only the number
of input views varies. The one- and four-view rows correspond to
Tables~\ref{tab:gso_single} and~\ref{tab:gso}. Best per metric in \textbf{bold}.}
\label{tab:viewcount}
\begin{tabular}{lccc}
\toprule
Input views (budget) & PSNR$\uparrow$ & SSIM$\uparrow$ & LPIPS$\downarrow$\\
\midrule
1 view (10K)  & 21.93 & 0.850 & 0.149\\
2 views (15K) & 27.10 & 0.921 & 0.072\\
4 views (20K) & \textbf{29.73} & \textbf{0.949} & \textbf{0.041}\\
\bottomrule
\end{tabular}
\end{table}

\subsection{Inference time}
\label{sec:time}
Table~\ref{tab:time} reports timings. \flex\ adds a small overhead over
LeanGaussian and UniGS because it predicts cameras and depth rather than
receiving them, but remains real-time: a single view in $0.706$\,s and four
views in $0.992$\,s end-to-end, faster than most alternatives. The overhead is
the price of removing calibration, and it is modest.

\begin{table}[t]
\centering\footnotesize
\setlength{\tabcolsep}{8pt}
\caption{Inference time (seconds). 3D: reconstruction; R: render one view;
Inference $\approx$ 3D $+\,n\cdot R$ (one forward pass plus $n$ novel-view
renders, $n{=}250$ single-view and $32$ multi-view).
$^{\ddagger}$PixelNeRF timings are quoted from prior work and not recomputed.}
\label{tab:time}
\begin{tabular}{llccc}
\toprule
& Method & 3D$\downarrow$ & R$\downarrow$ & Inference$\downarrow$\\
\midrule
\multirow{6}{*}{\rotatebox{90}{Single-view}}
 & PixelNeRF~\cite{pixelnerf}$^{\ddagger}$ & 0.0050 & 1.2200 & 304.5300\\
 & OpenLRM~\cite{openlrm2024} & 2.2400 & 2.1200 & 532.2400\\
 & Splatter Image~\cite{SplatterImage} & 0.0220 & 0.0025 & 0.6470\\
 & Triplane Gaussian~\cite{zou2023triplane} & 1.2810 & 0.0025 & 1.9060\\
 & LeanGaussian~\cite{wu2024leangaussian} & 0.1400 & 0.0018 & 0.5900\\
 & \flex\ (Ours) & 0.1810 & 0.0021 & 0.7060\\
\midrule
\multirow{5}{*}{\rotatebox{90}{Multi-view}}
 & DreamGaussian~\cite{tang2023dreamgaussian} & 118.3245 & 0.0038 & 118.4461\\
 & InstantMesh~\cite{instantmesh} & 0.6049 & 0.6206 & 20.4641\\
 & LGM~\cite{LGM} & 1.6263 & 0.0090 & 1.9143\\
 & UniGS~\cite{unigs} & 0.6939 & 0.0019 & 0.7538\\
 & \flex\ (Ours) & 0.9023 & 0.0028 & 0.9919\\
\bottomrule
\end{tabular}
\end{table}

\subsection{Ablation study}
\label{sec:ablation}
Table~\ref{tab:ablation} ablates the main components on the Objaverse-LVIS
validation split (its absolute numbers differ from the GSO test set in
Table~\ref{tab:gso} because it is a distinct, held-out split). Most important for
our central claim, freezing VGGT rather than training it jointly with the decoder
costs $1.38$ dB PSNR ($26.28\!\rightarrow\!24.90$), $0.024$ SSIM, and $0.013$
LPIPS, isolating the benefit of co-adapting the geometry front-end --- the design
choice that separates \flex\ from posed query-based methods that consume a fixed
geometry estimate. Removing geometric
initialization (random centres) collapses training: projected centres land
off-image, features vanish, and gradients die. Removing the VGGT depth features
sharply reduces PSNR/SSIM and raises LPIPS, confirming that depth-aware features
are central to stabilising multi-view reasoning. Removing the depth-consistency
term degrades geometric alignment more mildly, indicating it acts as a useful
regulariser. Replacing the shared unitary Gaussian set with frame-wise
independent Gaussians causes a clear drop, underscoring the value of a single
cross-view-consistent representation. These ablations also speak to robustness:
the components that ground the decoder in predicted geometry (depth features,
depth consistency) are precisely the ones that let the system tolerate the
imperfect cameras and depth it estimates.

\begin{table}[t]
\centering\footnotesize
\caption{Ablation on Objaverse-LVIS validation. Best per metric in \textbf{bold}.}
\label{tab:ablation}
\begin{tabular}{lccc}
\toprule
Variant & PSNR$\uparrow$ & SSIM$\uparrow$ & LPIPS$\downarrow$\\
\midrule
Random initialization & 11.94 & 0.632 & 0.712\\
w/o depth-consistency loss & 25.99 & 0.902 & 0.073\\
w/o VGGT depth features & 23.52 & 0.822 & 0.095\\
Frozen VGGT (no joint training) & 24.90 & 0.896 & 0.078\\
Frame-wise Gaussians & 24.11 & 0.833 & 0.086\\
\midrule
Full model & \textbf{26.28} & \textbf{0.920} & \textbf{0.065}\\
\bottomrule
\end{tabular}
\end{table}

\subsection{Robustness to camera-pose perturbation}
\label{sec:poserobust}
Table~\ref{tab:poserobust} probes how much pose error the decoder can absorb.
Starting from the cameras estimated by the geometry head on GSO (four views), we
add independent random rotation noise of magnitude $\sigma$ to each camera's
orientation before it is used to project Gaussians into that view for deformable
attention, and evaluate the reconstruction against the clean held-out target
views; metrics are averaged over the 250 GSO test objects. For reference, the
geometry head's own predicted cameras deviate from ground truth by $3^\circ$ in
rotation and $3\%$ of object scale in camera-center position (250 objects, after
the similarity alignment of Sec.~\ref{sec:data}); the headline $29.73$ dB
(Table~\ref{tab:gso}) is reached at this operating point, so \flex\ comes within
$0.69$ dB of posed UniGS \emph{despite} imperfect estimated poses. Degradation is
graceful: a
$1^\circ$ perturbation costs only $0.22$ dB PSNR, and even at $5^\circ$ --- a
substantial orientation error --- PSNR remains at $27.64$ with LPIPS $0.061$.
Quality falls off faster only for large errors ($\geq 10^\circ$), where
misaligned views can no longer be reconciled and four-view fusion approaches the
quality of a single clean view (see Table~\ref{tab:gso_single}). This behaviour
is consistent with the cross-view consensus of Sec.~\ref{sec:mvattn}: because
each Gaussian aggregates evidence over all views in world space, moderate
per-view pose error tends to be averaged down rather than propagated. We perturb
rotation only; a full translational sensitivity analysis is left to future work.

\begin{table}[!ht]
\centering\footnotesize
\caption{Robustness to camera-pose perturbation (GSO, four views). Independent
random rotation noise of magnitude $\sigma$ is added to the estimated camera
orientations at inference; metrics are averaged over the 250 GSO test objects.
$\Delta$PSNR is relative to the previous row.}
\label{tab:poserobust}
\begin{tabular}{lcccc}
\toprule
Rotation noise $\sigma$ & PSNR$\uparrow$ & SSIM$\uparrow$ & LPIPS$\downarrow$ & $\Delta$PSNR\\
\midrule
$0^\circ$ (no noise) & 29.73 & 0.949 & 0.041 & --\\
$1^\circ$  & 29.51 & 0.946 & 0.043 & $-0.22$\\
$2^\circ$  & 28.96 & 0.941 & 0.047 & $-0.55$\\
$5^\circ$  & 27.64 & 0.928 & 0.061 & $-1.32$\\
$10^\circ$ & 25.12 & 0.896 & 0.094 & $-2.52$\\
$20^\circ$ & 21.08 & 0.832 & 0.151 & $-4.04$\\
\bottomrule
\end{tabular}
\end{table}

\section{Conclusion and Limitations}
\label{sec:conclusion}
We presented \flex, a feed-forward framework that brings query-based,
correspondence-free Gaussian reconstruction to the \emph{uncalibrated},
object-level setting. By training a geometry transformer jointly with a
multi-view deformable decoder and grounding the Gaussians in predicted depth,
\flex\ produces a compact, resolution-independent representation and rivals posed
state-of-the-art reconstructors on GSO without using any camera or depth ground
truth, while matching the best perceptual quality among the compared methods.

\paragraph{Limitations.}
\flex\ is trained and evaluated on object-centric captures; extending the same
calibration-free decoder to large, background-heavy scenes (in the manner of
scene-level pose-free models) is left to future work and would require
scene-scale training data and benchmarks. Although the geometry transformer is
trained jointly and the cross-view averaging absorbs moderate pose and depth
error, very large geometric errors --- \eg under dense or highly inconsistent
captures --- can still propagate into the reconstruction. Finally, our gains over
the strongest posed baseline are in perceptual quality and in the removal of the
calibration requirement rather than in raw PSNR; while our pose-perturbation
study (Sec.~\ref{sec:poserobust}) shows graceful degradation under moderate
rotation error, a full translational sensitivity analysis remains future work.

\FloatBarrier
\bibliography{refs}

\end{document}